# Forecasting, Causality, and Impulse Response with Neural Vector Autoregressions

**Kurt Izak Cabanilla**

Thinking Machines Data Science

**Kevin Thomas Go**

Thinking Machines Data Science

## Abstract

Incorporating nonlinearity is paramount to predicting a dynamical system's future states, its response to shocks, and its underlying causal network. However, most existing methods for causality detection and impulse response, such as Vector Autoregression (VAR), assume linearity and are thus unable to capture the complexity. Here, we introduce a vector autoencoder nonlinear autoregression neural network (VANAR) capable of both automatic time series feature extraction for its inputs and functional form estimation. We evaluate VANAR in three ways: first in terms of pure forecast accuracy, second in terms of detecting the correct causality between variables, and lastly in terms of impulse response where we model trajectories given external shocks. These tests were performed on a simulated nonlinear chaotic system and an empirical system using Philippine macroeconomic data. Results show that VANAR significantly outperforms VAR in the forecast and causality tests. VANAR has consistently superior accuracy even over state of the art models such as SARIMA and TBATS. For the impulse response test, both models fail to predict the shocked trajectories of the nonlinear chaotic system. VANAR was robust in its ability to model a wide variety of dynamics, from chaotic, high noise, and low data environments to macroeconomic systems.

# Introduction

Many dynamical systems such as those in neuroscience, ecology, epidemiology, and economics can be highly nonlinear [3]. For example in macroeconomics, the entire economy is a complex network comprised of various industries and sectors, interest rates, exchange rates, inflation rates, labor, education, and household consumption among others [12]. These components have causal interactions with each other that can evolve. Introducing a shock such as a sudden increase in money supply from the central bank or fiscal investments from the government can drastically change the entire economy especially in the short run. How can we better know the impact of monetary policy over time? How much do we need to decrease the interest rate in order to improve economic output without too much inflation? Accounting for the complex relationships between variables in a system are thus important in forecasting the future state of the system, especially with respect to the effect of shocks.

Classical approaches such as Vector Autoregression assume linearity and have produced unsatisfactory forecasts and simulations especially in macroeconomics [11]. Furthermore, even standard nonlinear forms of VAR such as Threshold, Markov Switching, and Time Varying Parameters not only assume the type of nonlinearity but are also quite ad hoc in their specifications and hence require considerable tuning for different cases [7].

Various machine learning methods have shown promise in time series analysis due to their ability to model more complex, non-linear relationships. However, applications have primarily been on pure forecasting [2] [9], and to our knowledge there is little work on causality analysis and impulse response estimation. Neural network approaches to VAR have been very sparse with the focus mostly on forecasting and only recently on causality [5]. However, these neural VAR models only take in pure time series lags as input. They do not utilize other time series features which have been known to increase forecast accuracy in other models [9]. While manually derived features are feasible and powerful for univariate forecasting, in dynamical systems the number of important features can increase exponentially due to interaction terms. In addition, expert knowledge is needed when engineering features, as the features considered "important" (as well as their functional relationships) vary from system to system. This makes feature engineering for multivariate time series models highly untenable, arbitrary, and time consuming.

To address these limitations, we propose the vector autoencoder nonlinear autoregression (VANAR). It is a combination of two neural networks--an autoencoder for automatically extracting time series features for input processing, and a neural multilayer perceptron vector autoregression for automatically approximating the best functional form to fit the dynamical system.

We assess the performance of VANAR (and compare it against existing forecasting methods) through four tests which we describe briefly here and expound on in the Methods section. First, we conduct a simulated forecast test, where we compare long horizon n-step ahead forecasts. Second, we test VANAR's ability to correctly detect causal relationships between variables. Third, we examine VANAR's ability to accurately simulate the dynamic effects of a shock through time. For these first three tests, we assess VANAR's performance on a simulated system (a two variable chaotic time series dataset simulated from a coupled logistic dynamical system) and one real-world system (on Philippine macroeconomic data). We also compare the performance between VANAR and VAR--a popularly-used method for causality detection and impulse response analysis for multivariate time series. Finally, we conduct one step ahead forecast tests using univariate VANAR, and compare it against several state-of-the-art statistical and machine learning models. Specifically, we compare it against the Seasonalized Autoregressive Integrated Moving Average (SARIMA), Exponential Smoothing State Space Model With Box-Cox Transformation,

ARMA Errors, Trend And Seasonal Components (TBATS) [4], and an autoregressive single hidden layer multilayer perceptron (MLP).

Although we could test the accuracy of n-step ahead forecasts for non-simulated time series, we cannot do the same for impulse response precisely because we do not have access to true counterfactuals. In real life, there is only a single actualization. Using simulated data generated from a given dynamical system, however, we can create the true counterfactuals and thus gauge the accuracy of the impulse response functions. Nonetheless, we still do an impulse response analysis for the macroeconomic empirical dataset.

## Methods

The methods section will be structured as follows. First, we will describe classic VAR to introduce one approach of multivariate autoregressive time series analysis. We then describe VANAR and how it relates to (and builds on) VAR-like methods while utilizing feature extraction and functional form estimation capabilities of deep learning (this includes a section on the different possible architectures of VANAR, and how the optimal lags of VANAR are selected). In the third section, we elaborate on our tests of VANAR's performance: (1) forecast accuracy, (2) causality detection, (3) impulse response analysis, and (4) one step ahead univariate forecasts on empirical data. In the final section, we describe the datasets we conduct tests on, and how the simulated datasets were generated.

For discussing the first three Methods sections, let us consider a toy dynamic system with only two variables (but it can be easily extended for *N* variables):

Let the information set $\Omega_T = \{(x_1, y_1); (x_2, y_2); \ldots (x_t, y_t) \ldots (x_T, y_T)\}$ be two time series of length $T$ and generated by some underlying discrete dynamical system $M$ and initial conditions $X_o$. $\Omega_T$ is essentially a sample trajectory $S(t \mid X_o)$. Let

$$\hat{x}_t = \hat{f}\left(x_{t-1}, x_{t-2}, \ldots x_{t-p}, y_{t-1}, y_{t-2}, \ldots y_{t-p}\right)$$

$$\hat{y}_t = \hat{g}(x_{t-1}, x_{t-2}, \ldots x_{t-p}, y_{t-1}, y_{t-2}, \ldots y_{t-p})$$

be an estimated system of difference equations approximating $M$ given the information set, or $E[M \mid \Omega_T] = \hat{F}$, where $p$ is the max lag and $\hat{F} = (\hat{f}, \hat{g})'$.

This estimated dynamical system can be represented in vector form as

$$\hat{X}_t = \hat{F}\left(X_{t-1}, X_{t-2}, \ldots X_{t-p}\right) \text{ where } X_t = (x_t, y_t)' \text{ and } \hat{X}_t = (\hat{x}_t, \hat{y}_t)'.$$

## Vector Autoregression

In a VAR with $p$ lags or a VAR-p, the estimated dynamic system has a linear functional form:

$$\hat{X}_t = \hat{F}\left(X_{t-1}, X_{t-2}, \ldots X_{t-p}\right) = \Phi_1 X_{t-1} + \Phi_2 X_{t-1} \ldots + \Phi_p X_{t-p} =$$

$$E[\Phi_1 X_{t-1} + \Phi_2 X_{t-1} \ldots + \Phi_p X_{t-p} + \xi_t]$$

Where $\Phi_i$ are 2x2 (or $N \times N$ if there N variables in the system) coefficient matrices to be estimated and $\xi_t$ is a vector standard normal error term that has no serial autocorrelation: $E\left[\xi_t \xi_{t-1}\right] = 0$ and $E\left[\xi_t\right] = 0$

We use ordinary least squares (OLS) to estimate VAR and the classical Akaike Information Criterion (AIC) to select the optimal lag $p$.

## Vector Autoencoder Nonlinear Autoregression

A Vector Autoencoder Nonlinear Autoregression with *p* lags, or a VANAR-p, model has two components: (1) an autoencoder for the input processing; and (2) a vector autoregressive neural network for the dynamical system estimation.

The autoencoder is comprised of two multilayer perceptrons neural networks (MLP): an encoder that compresses its input data to a lower dimension, and a decoder that takes the compressed output of the encoder and decompresses it to approximate the data in its original dimension. The objective of the encoder is to extract the most important features of its input data, while the objective of the decoder is to reconstruct the original inputs based on those important features alone. In the context of VANAR, the encoder serves as the "automatic" feature extractor (specifically, by reducing the dimensionality of lagged variables in the time series), while the decoder is a means to measure how well these reduced features approximate the original features.

The encoder $E$ takes in the vectors of the form $\left(X_{t-1}, X_{t-2}, \ldots X_{t-p}\right)$ comprised of the *p* lags of two variables and outputs a lower dimensional vector $R^{embedding\ dimension}$. Given the two variable system, the dimension of this input vector is thus *p x 2,* while the output is simply the user selected *embedding dimension.*

$$E : R^{p\ x\ 2} \to R^{embedding\ dimension}$$

The decoder *D* then takes in the output of the encoder and outputs back a vector as close as possible to the encoder input vector:

$$D : E(R^{p\ x\ 2}) \to R^{p\ x\ 2}$$

$$Autoencoder = D \circ E \ : \ R^{p\ x\ 2} \to R^{embedding\ dimension} \to R^{p\ x\ 2}$$

The output of the encoder $E(X_{t-1}, X_{t-2}, \ldots X_{t-p})$ is essentially the vector of automatically extracted time series features. This is then concatenated to $(X_{t-1}, X_{t-2}, \ldots X_{t-p})$ (i.e., the original vectors in a classic VAR) and used as input to another multilayer perceptron, $\hat{N}$

, which is basically a VAR modified such that each equation in the VAR is a multilayer perceptron. it is a nonlinear vector autoregression. $\hat{N}$ without an autoencoder is simply $\hat{N}: R^{p \times 2} \to R^2$. With an activated autoencoder, $\hat{N}$ is $\hat{N}: R^{(p+embedding\ dimension) \times 2} \to R^2$.

It is important to note that $\hat{N}$ need not necessarily be a neural network, it can be any other form of estimator whether linear or nonlinear such as a random forest regressor or whatnot. For the purposes of this paper, however, the nonlinear vector autoregression is a multilayer perceptron neural network.

For a VANAR-p with two hidden layers and *deactivated autoencoder*, the first equation $\widehat{x_t}$ of $\hat{N}$ is represented as:

$$\widehat{x_t} = \hat{f}(x_{t-1}, x_{t-2}, \ldots x_{t-p}, y_{t-1}, y_{t-2}, \ldots y_{t-p}) = W_3^T \cdot h_2 (W_2 \bullet h_2 ( W_1 \cdot x + b_0) + b_1)$$

Where $x = (X_{t-1}, X_{t-2}, \ldots X_{t-p})'$, $W_1$ denotes the weight matrix consisting of the values of $w_{ij}^1$ with $i$ as row and $j$ as column coordinates, $b_0$ is a vector whose elements are all $b_0$, $h^1$ is a vector valued function where the sigmoid, or any other function (called activation functions) such as the tanh, is applied to the vector $W_1 x + b_0$, $W_2$ denotes the weight matrix consisting of the values of $w_{kj}^2$, and so on until the vector $h^2$ is multiplied by the transpose of the weight vector $W_3$ to produce the scalar output $\hat{x}_t$. The weights of VANAR are then iteratively estimated using a gradient descent based optimization algorithm.

The estimated dynamical system for **M** given a VANAR-p with an activated autoencoder is the following:

$$\hat{X}_t = \hat{F}\left(X_{t-1}, X_{t-2}, \ldots X_{t-p}\right) = \hat{N}\left(X_{t-1}, X_{t-2}, \ldots X_{t-p}, E(X_{t-1}, X_{t-2}, \ldots X_{t-p})\right)$$

$\hat{F}$ is once more comprised of two different equations, one for each variable:

$$\hat{x}_t = \hat{f}(x_{t-1}, x_{t-2}, \ldots x_{t-p}, y_{t-1}, y_{t-2}, \ldots y_{t-p}, E\left(x_{t-1}, x_{t-2}, \ldots x_{t-p}, y_{t-1}, y_{t-2}, \ldots y_{t-p}\right))$$

$$\hat{y}_t = \hat{g}(x_{t-1}, x_{t-2}, \ldots x_{t-p}, y_{t-1}, y_{t-2}, \ldots y_{t-p}, E\left(x_{t-1}, x_{t-2}, \ldots x_{t-p}, y_{t-1}, y_{t-2}, \ldots y_{t-p}\right))$$

A VANAR-p with two hidden layers, the first equation $\hat{x}_t$ in a VANAR-p can be represented as:

$$\hat{x}_t = W_3^T \cdot h_2 (W_2 \bullet h_2 ( W_1 \cdot x + b_0) + b_1)$$

Where $x = (X_{t-1}, X_{t-2}, \ldots X_{t-p}, E\left(X_{t-1}, X_{t-2}, \ldots X_{t-p}\right))'$

The univariate version of VANAR is the Autoencoder Nonlinear Autoregression, or ANA. It is simply:

$$\hat{x}_t = \hat{f}(x_{t-1}, x_{t-2}, \ldots x_{t-p}, E\left(x_{t-1}, x_{t-2}, \ldots x_{t-p}\right))$$

$$\hat{y}_t = \hat{g}(x_{t-1}, x_{t-2}, \ldots x_{t-p}, E\left(x_{t-1}, x_{t-2}, \ldots x_{t-p}\right))$$

**VANAR Architecture and Lag Selection**

The number of lags *p* is decided by an initial VAR estimation using the classical Akaike Information Criterion (AIC), thus VANAR has the same *p* as VAR. VANAR has an architecture that is shallow but extremely wide. It only has two hidden layers but the number of neurons in each layer often ranges from 3000 to 5000. The optimizer is the AdaGrad algorithm with a learning rate of 0.0001. The activation functions are RELU for the hidden layers and linear for the output layer. The autoencoder consists of 3 hidden layers each with RELU activations and a linear activation for the output layer. Note that the autoencoder can be activated or deactivated depending on the length of the optimal lag and the validation set forecast accuracy. If the ideal lag is below 4, then feature extraction will not make much sense and thus is no longer considered. ANA has the same architecture and two forms as VANAR but instead is comprised of a single equation as it is univariate. The architecture diagram of a VANAR-p is shown below:

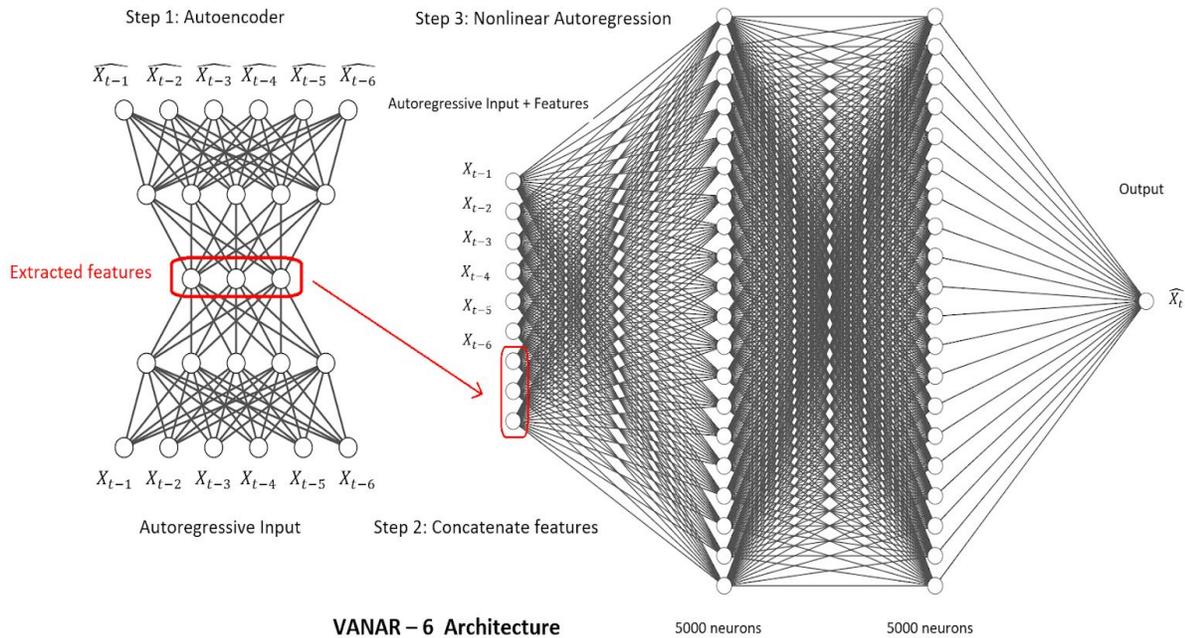

**Fig 1.** VANAR Architecture

## Assessment of VANAR performance

To assess the performance of VANAR, we conduct four tests: (1) an n-step ahead forecast test, (2) a causality test, (3) an impulse response test, and (4) an empirical one step ahead univariate forecast test. The error metric used is the classical Root Mean Squared Error (RMSE); a lower RMSE means better forecast accuracy. For the first three tests, VANAR and VAR will be compared, while in the fourth test ANA (i.e., univariate VANAR) will be compared against other univariate forecasting models.

**Forecasting Accuracy**

The first test is the n-step ahead forecast comparison between VAR and VANAR, where RMSE of the two methods are compared. We do this for a long horizon forecast (a twenty step ahead forecast) and a short horizon forecast (a ten step ahead forecast).

The method that results in a lower RMSE is considered to be more accurate in forecasting across that time horizon.

This initial test is important, because the causal inference and impulse response predictions of a model may not be robust or consistent if the model cannot make sufficiently accurate predictions relative to its competitors. In addition, many time series analysis applications in business and industry focus on forecasting accuracy [10]; thus, determining how well VANAR performs this function (and how it compares to VAR) would be of great interest to potential users of VANAR for its forecasting ability.

**Granger Causality**

The second test is the Granger Causality Test. Given the forecast scores of VANAR and VAR from the first test, we evaluate the forecast scores of their univariate forms--ANA and AR, respectively. We then compare RMSE according to the definition of Granger Causality: Granger Causality states that variable X Granger Causes variable Y if the history of X improves the forecast accuracy of a model that only uses the history of Y (see Definition 1, below). Thus, if the RMSE of VANAR (including X and Y) is lower than that of ANA (of Y only), this suggests that information encoded in lags of X is important in forecasting future states of Y, and X is thought to Granger Cause Y. The outputs of this test are binary: either the model correctly identifies the proper causal relationship or not.

We use the definition of Granger Causality in Hamilton (1996) but extend it to nonlinear functions:

**Definition 1.** *Given a dynamical system estimate* $\hat{F}$, *out of sample values* $y_t$, *predicted values* $\hat{y}_t$, *and an error metric* $L(\hat{y}_t, y_t)$ *such as the Root Mean Squared Error (RMSE), a variable x is said to not Granger Cause variable y if the test set (out of sample) error*

$$L(\hat{g}(x_{t-1}, x_{t-2}, \ldots x_{t-p}, y_{t-1}, y_{t-2}, \ldots y_{t-p}), y_t)$$

*Is greater than or equal to*

$$L(\hat{g}(y_{t-1}, y_{t-2}, \ldots y_{t-p}), y_t)$$

*If otherwise, x is said to Granger Cause y.*

*Alternatively, variable x is said to Granger Cause variable y if*

$$1 - \frac{L(\hat{g}(x_{t-1}, x_{t-2}, \ldots x_{t-p}, y_{t-1}, y_{t-2}, \ldots y_{t-p}), y_t)}{L(\hat{g}(y_{t-1}, y_{t-2}, \ldots y_{t-p}), y_t)} > 0$$

*We denote the expression above as the **causality score**. The closer it is to 1, the stronger the evidence for a causal effect.*

The test set forecasts are used rather than in sample predictions because train set overfitting can lead to misleading results.

It follows that the degree or strength of the Granger Causal variables can be quantified by how much they improve the error of a purely univariate model. For instance, in a system with three variables (*x, y,* and *z*), variable *x* has a stronger Granger Causal effect on *y* than variable *z* if

$$L(\hat{g}(x_{t-1}, x_{t-2}, \ldots x_{t-p}, y_{t-1}, y_{t-2}, \ldots y_{t-p}), y_t)$$

Is less than

$$L(\hat{g}(z, z_{t-2}, \ldots z_{t-p}, y_{t-1}, y_{t-2}, \ldots y_{t-p}), y_t)$$

Is less than

$$L(\hat{g}(y_{t-1}, y_{t-2}, \ldots y_{t-p}), y_t)$$

This nonlinear Granger Causality need not be restricted to bivariate or pairwise testing. Comparing the accuracy of an *n* variable model with a univariate model can show how a collection of *n-1* variables can have a causal effect on the *nth* variable, we call this *n-1* Granger Causality.

**Impulse Response Analysis**

The impulse response test is also a forecast comparison like the first test but instead with an alternate data input due to the impulse shock. The model forecasts are then the estimated impulse response paths given the shock. This is then compared with the simulated true impulse response paths given the underlying dynamical system.

We define a general impulse response similar to (Kilian & Lutkepohl, 2017). Recall the system we defined at the beginning of the Methods section, where the information set $\Omega_T = \{(x_1, y_1); (x_2, y_2); \ldots (x_t, y_t) \ldots (x_T, y_T)\}$ be two time series of length *T* and generated by some underlying discrete dynamical system $M$ and initial conditions $X_o$. This estimated dynamical system can be represented in vector form as

$\hat{X}_t = \hat{F}\left(X_{t-1}, X_{t-2}, \ldots X_{t-p}\right)$ where $X_t = (x_t, y_t)'$ and $\hat{X}_t = (\hat{x}_t, \hat{y}_t)'$.

How will the system evolve given a sudden increase in one of the variables at a certain period? For instance, if *x* is shocked at time *T* by an impulse $\varepsilon$, then what is the new trajectory of both *x* and *y* over time? Using this example, we will have a new information set representing a counterfactual which we denote as:

$\Omega_{T,x}^{\varepsilon} = \{(x_1, y_1); (x_2, y_2); \ldots (x_t, y_t); \ldots (x_T + \varepsilon, y_T)\}$.

Generally, $\Omega_{T,x}^{\varepsilon}$ is defined as the **impulse information set** where we have $T$ as the period of the shock, $x$ as the shocked variable, and $\varepsilon$ as the magnitude of the shock. We use this new information set as the input to the estimated dynamical system. We recursively produce outputs for all the variables in the system until the designed time path length $H$ has been reached. The resulting set of values, or the counterfactual time path, is called the impulse response function. To obtain the predictions for *T+1* we have:

$$\hat{x}_{T+1} = \hat{f}\left(x_T + \varepsilon, x_{T-1}, \ldots x_{T-p}, y_T, y_{T-1}, \ldots y_{T-m}\right) = \hat{f}(A_T \subseteq \Omega_{T,x}^{\varepsilon})$$

$$\hat{y}_{T+1} = \hat{g}\left(x_T + \varepsilon, x_{T-1}, \ldots x_{T-p}, y_T, y_{T-1}, \ldots y_{T-m}\right) = \hat{g}(B_T \subseteq \Omega_{T,x}^{\varepsilon})$$

We update the impulse information set for the next period, $\Omega_{T+1,x}^{\varepsilon}$, with the predictions $\hat{X}_t$ in order to track the dynamic effects of the initial shock:

$$\Omega_{T+1}^{\varepsilon} = \{(x_1, y_1); (x_2, y_2); \ldots (x_t, y_t) \ldots (x_T + \varepsilon, y_T), (\hat{x}_{T+1}, \hat{y}_{T+1})\}$$

Thus we have the new state for *T+2*:

$$\hat{x}_{T+2} = \hat{f}\left(\hat{x}_{T+1}, x_T + \varepsilon, \ldots x_{T-p+1}, \hat{y}_{T+1}, y_T, \ldots y_{T-m+1}\right) = \hat{f}(A_{T+1} \subseteq \Omega_{T+1,x}^{\varepsilon})$$

$$\hat{y}_{T+2} = \hat{g}\left(\hat{x}_{T+1}, x_T + \varepsilon, \ldots x_{T-p+1}, \hat{y}_{T+1}, y_T, \ldots y_{T-m+1}\right) = \hat{g}(A_{T+1} \subseteq \Omega_{T+1,x}^{\varepsilon})$$

And so on until $\Omega_{T+H,x}^{\varepsilon}$ is recursively generated where *H* is the length of the time path after the shock period. $\Omega_{T+H,x}^{\varepsilon}$ is defined as the **impulse information set recursively generated by** $\hat{F}$.

The predicted values of the entire system after the period of the shock is then simply all the elements of $\Omega_{T+H,x}^{\varepsilon}$ for $t > T$. We define this formally.

**Definition 2 (Impulse Path):** *Given an impulse information set $\Omega_{T+H,x}^{\varepsilon}$ recursively generated by an estimation $\hat{F}$ of $\Omega_T$, the **impulse path** of $\hat{F}$ given a shock $\varepsilon$ on variable x at time T is:* $R_x^{\varepsilon}(t) := (\hat{x}_t, \hat{y}_t) \in \Omega_{T+H}^{\varepsilon}$ *for* $t > T$

The impulse response is then simply the impulse path minus the the impulse path given no shock:

**Definition 3 (Impulse Response):** *Given an impulse information set $\Omega_{T+H,x}^{\varepsilon}$ recursively generated by an estimation $\hat{F}$ of $\Omega_T$, the impulse response of $\hat{F}$ given a shock $\varepsilon$ on variable $x$ at time $T$ is:* $I_x^{\varepsilon}(t) := R_x^{\varepsilon}(t) - R_x^0(t)$ *for* $t > T$

To assess the performance of VANAR (and VAR) in predicting the effects of a shock on the system, we compare the VANAR-estimated (VAR-estimated) impulse response $I_x^{\varepsilon}(t)_{VANAR}$ ($I_x^{\varepsilon}(t)_{VAR}$) against the *true* impulse response of the system $I_x^{\varepsilon}(t)_{True}$. Clearly, this comparison can only be made for simulated systems, where we have access to true counterfactuals.

**Univariate One Step Ahead Forecasting Accuracy**

The fourth and last test is an empirical one step ahead forecast test, which will only be conducted on empirical Philippine Macroeconomic data. Several models will be compared on a one step ahead forecast accuracy test using Philippine monthly inflation rate and tourist arrivals. The models will be a univariate VANAR, the Seasonalized Autoregressive Integrated Moving Average (SARIMA), Exponential Smoothing State Space Model With Box-Cox Transformation, ARMA Errors, Trend And Seasonal Components (TBATS) [4], and an autoregressive single hidden layer multilayer perceptron (MLP).

Training set for inflation is from February 1990 to December 2017. The test set is the entirety of 2018 plus January of 2019, a total of thirteen observations. For tourist arrivals, the training set is from January 1991 to December 2016. The test set is the entirety of 2017 and 2018, a total of twenty four observations.

## Datasets for Analysis

We assessed the performance of VANAR (and the other aforementioned forecasting models) on two datasets: a simulated two variable chaotic system, and an empirical

Philippine macroeconomic dataset to test its performance on real-world data. Note that Granger Causality and Impulse Response Analysis can only be properly evaluated on the two simulated systems due to access to "true" system dynamics.

For the empirical dataset, only the forecast tests (tests 1 and 4) can be properly evaluated since we do not know the underlying causal dynamical system (i.e., cannot check correctness of Granger causality test) and thus have no access to the real counterfactuals (i.e., cannot compare impulse responses between VAR- and VANAR-estimated trajectories and the "true" trajectory given a shock). Nonetheless, causality analysis as well as impulse response analysis will still be done for insights into the Philippine economy.

**Nonlinear System with Two Variables (System 1)**

We simulated a two-variable chaotic system following [13]:

$$X_t = X_{t-1}\left[3.8 - 3.8\, X_{t-1} - 0.02\, Y_{t-1}\right] + e_t \quad (1)$$
$$Y_t = Y_{t-1}\left[3.5 - 3.5\, Y_{t-1} - 0.1\, X_{t-1}\right] + e_t \quad (2)$$

where X and Y are coupled variables, and $e_t$ is noise. For this paper, we refer to (1) and (2) as System 1. For the first scenario, we simulate the entire system without any noise (i.e., $e_t = 0$). This system produces the so-called 'mirage correlation', where two variables can appear to be correlated or anti-correlated at certain points in time, then lose this pattern at other points in time (**Fig 2**). However, when considering the entire time series of n = 1000 steps, the Spearman correlation between the two variables is only 0.09, despite them being coupled.

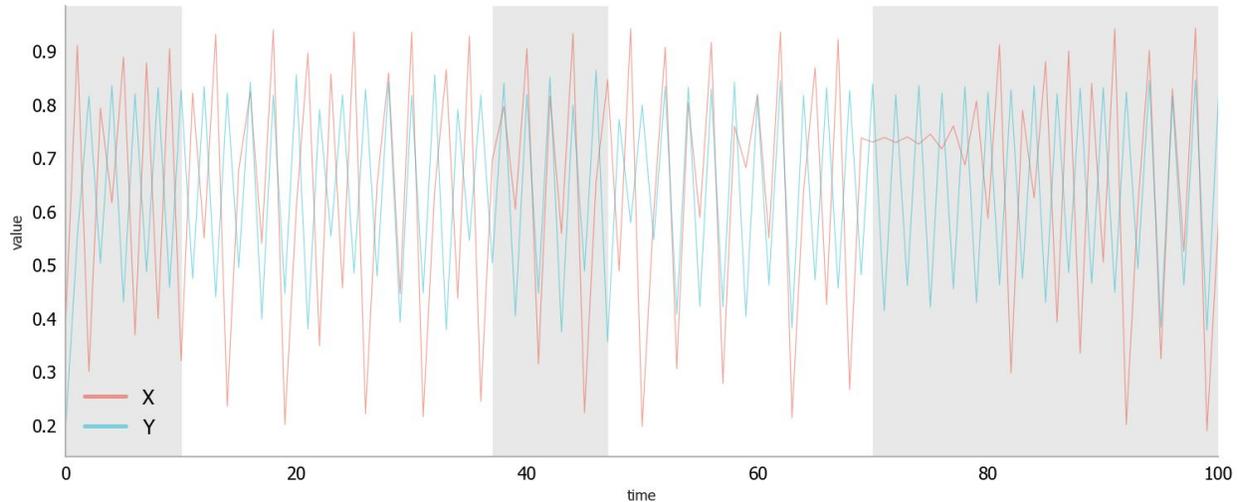

**Fig 2.** Illustration of 'mirage correlation' in System 1. Depending on the slice of the time series being examined, the two variables in the system can appear anti-correlated (first greyed section), correlated (second greyed section), or no clear patterns (third greyed section). Spearman's correlation of entire time series (n = 1000 steps) is 0.09.

For this system, we have four scenarios: the first is the Default Scenario which has no noise and has the exact functional form as (1) and (2). The second is the No Interaction scenario where the interaction coefficients of (1) and (2) are zero which implies that X has no effect on Y and vice versa (this is to test VANAR's precision). The third scenario is Noise 1 which is (1) and (2) but with a white noise error term with a mean of zero and a standard deviation of 0.1 (this is to test VANAR's sensitivity to high noise). and the fourth is Noise 2 which is the same as the previous scenario but instead with a standard deviation of 0.01 (test of VANAR's sensitivity to low noise). Note that when the noise is added, it is not part of the recursion but rather is added only once the system has been fully simulated. Hence it only distorts the pattern and is considered as "observation error".

Furthermore, for each scenario, we conduct the tests across three environments varying in the amount of data provided: a High Data Richness Environment where there are 850 observations available for training, a Medium Data Richness Environment where there are 350 observations available for training, and a Low Data Richness Environment where there are only 50 observations available for training. This is done to test VANAR's sensitivity to training set size. For all scenarios, the test set is the next 20 steps from the end of the training set (see Appendix for full dataset).

**Philippine Macroeconomic Data (System 2)**

To assess VANAR's performance on real-world data, we used VANAR (and VAR) to model the Philippine economy using the following macroeconomic variables: GDP annual growth rate, inflation rate, employment rate, interest rate, lending rate, M1 money supply (log transformed), fiscal expenditure (log transformed), remittances (log transformed), and industrial production. Despite some of the variables being nonstationary, we did not difference any of them in order to keep the full information of the dynamics necessary for the impulse response analysis. The data was quarterly, ranging from 1992 to all of 2018. Monthly variables were aggregated to mean quarterly frequency. We refer to the macroeconomic dataset as System 2.

The test set is comprised of the last four quarters which is the entire year of 2018. We first compared the forecast accuracy of the models with respect to GDP growth and then we analyzed the dynamic effects of expansionary fiscal and monetary policy shocks on the economy.

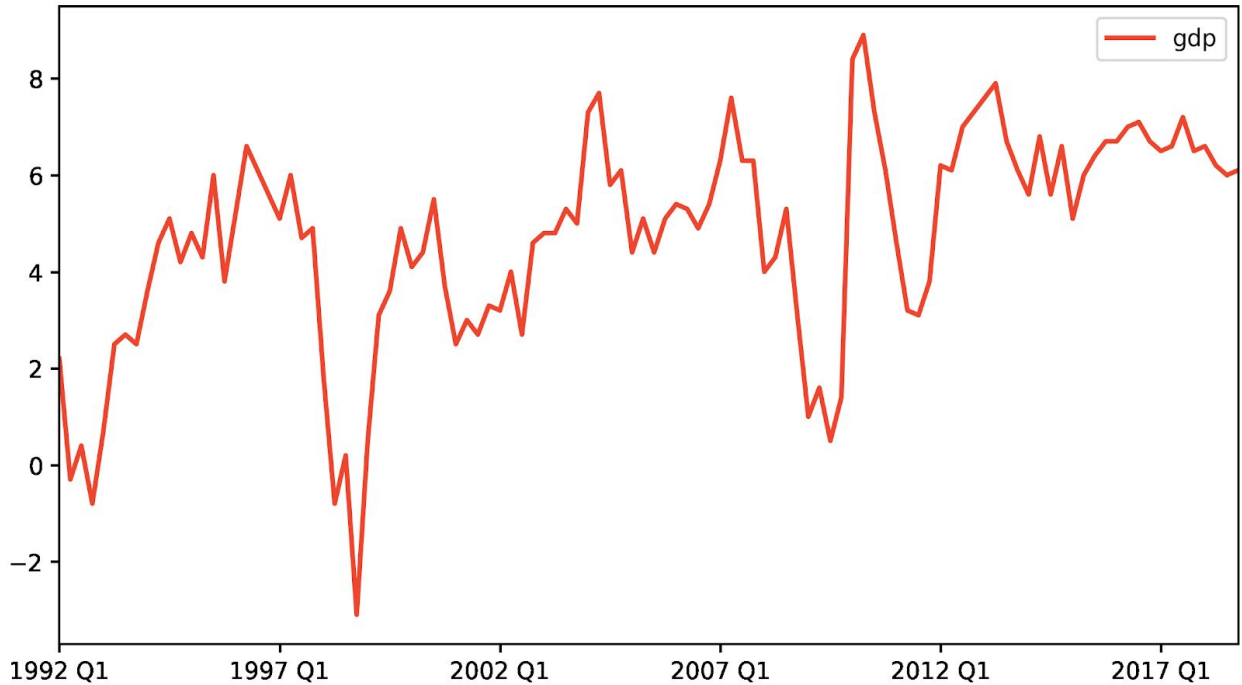

**Fig 3.** Philippine GDP Year on Year Growth, quarterly frequency.

# Results

## Nonlinear System with Two Variables

### Forecasting Accuracy

For all scenarios and for all environments, VANAR outperformed VAR for the long horizon forecast test (20 steps ahead) and significantly outperformed VAR in the short horizon test, often being twice as accurate for the short horizon (see Appendix for detailed results of all scenarios). An active autoencoder VANAR rather than a deactivated autoencoder was used for the vast majority of scenarios and environments since it yielded improved accuracy.

**Granger Causality**

For all scenarios and for all environments except one, VANAR detected the correct causality, while VAR failed in several scenarios and environments (**Table 1**). VANAR failed to detect causality only in the high noise environment (i.e., Noise 1), suggesting that caution must be exercised when attempting to infer causality from data with low signal-noise ratios (in our simulated system, values of *x* and *y* range from [0,1], and noise standard deviation was = 0.1). Nevertheless, VANAR still correctly detected causality more often than VAR, and was able to correctly detect causality in the low noise environment (i.e., Noise 2).

**Table 1.** Causality tests of VANAR and VAR across different scenarios and data richness environments. If causality is incorrectly detected, details on which variables detection failed is given in parentheses.

| Scenario | Method | High Data Richness | Medium Data Richness | Low Data Richness |
|---|---|---|---|---|
| **Default** | **VANAR** | Correct | Correct | Correct |
| | **VAR** | Incorrect (both) | Correct | Incorrect (fails to detect y on x) |
| **No Interaction** | **VANAR** | Correct | Correct | Correct |
| | **VAR** | Correct | Correct | Correct |
| **Noise 1** | **VANAR** | Incorrect (fails to detect x on y) | Correct | Correct |
| | **VAR** | Incorrect (both) | Correct | Incorrect (fails to detect x on y) |

| Noise 2 | VANAR | Correct | Correct | Correct |
| | VAR | Incorrect (both) | Correct | Incorrect (fails to detect y on x) |

**Impulse Response**

The impulse response test was done only in the default system scenario and the rich data environment. A shock of 0.1 was added to variable *y* at time step 850 and we forecasted the impulse response of *x* for twenty periods given the shock.

Results showed that the impulse response functions of both VAR and VANAR remained very close to zero at all times while the true impulse response remained close to zero only for around the first ten time steps and diverged drastically from there. This demonstrates the sensitivity to initial conditions that are characteristic of chaotic systems [1]. Furthermore, it implies that given a chaotic system and a single trajectory generated from initial conditions, models estimated on that trajectory tend to be dynamically stable. This can perhaps be solved by adding a noise term inside the true system itself and generating a random trajectory given the same initial conditions. Models estimated on that random trajectory might learn how the system reacts to shocks. Theoretical justifications for this phenomena are a subject of future research.

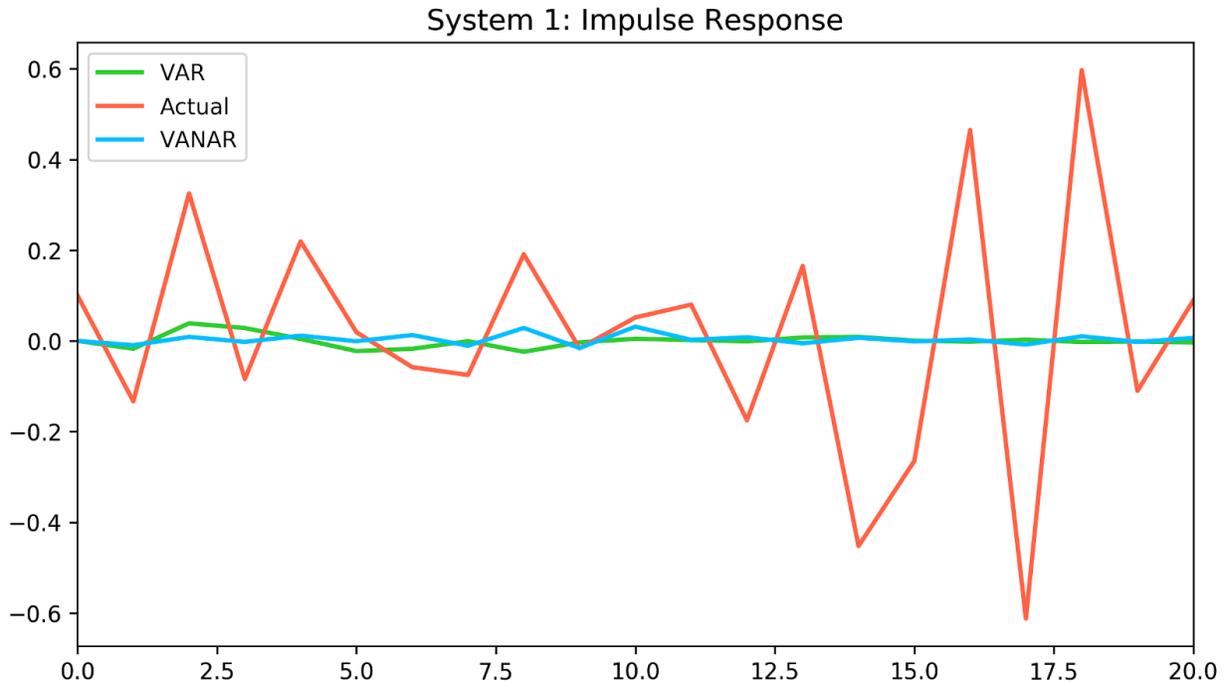

**Fig 4.** Impulse response forecasts of VAR, VANAR, and System 1 for variable *x* response to a shock of 0.1 added to variable *y* at time step 850.

## Philippine Macroeconomy

### Forecasting Accuracy

VANAR significantly outperformed VAR in the four steps ahead forecast test (VANAR RMSE = 0.336; VANAR RMSE = 0.098). The VAR selected here was the best possible VAR by iterating across different lag orders and deterministic exogenous components over the test set itself. The optimal VAR was a VAR-4 with both a constant and a trend component; its input used all the variables described in the Methods section. To determine the optimal VANAR, a pairwise causality analysis was done for the GDP variable to detect all variables that Granger cause GDP growth. The optimal VANAR was a VANAR-4 with activated feature extraction that used only M1 Money Supply, Fiscal Expenditure, and Remittances as inputs.

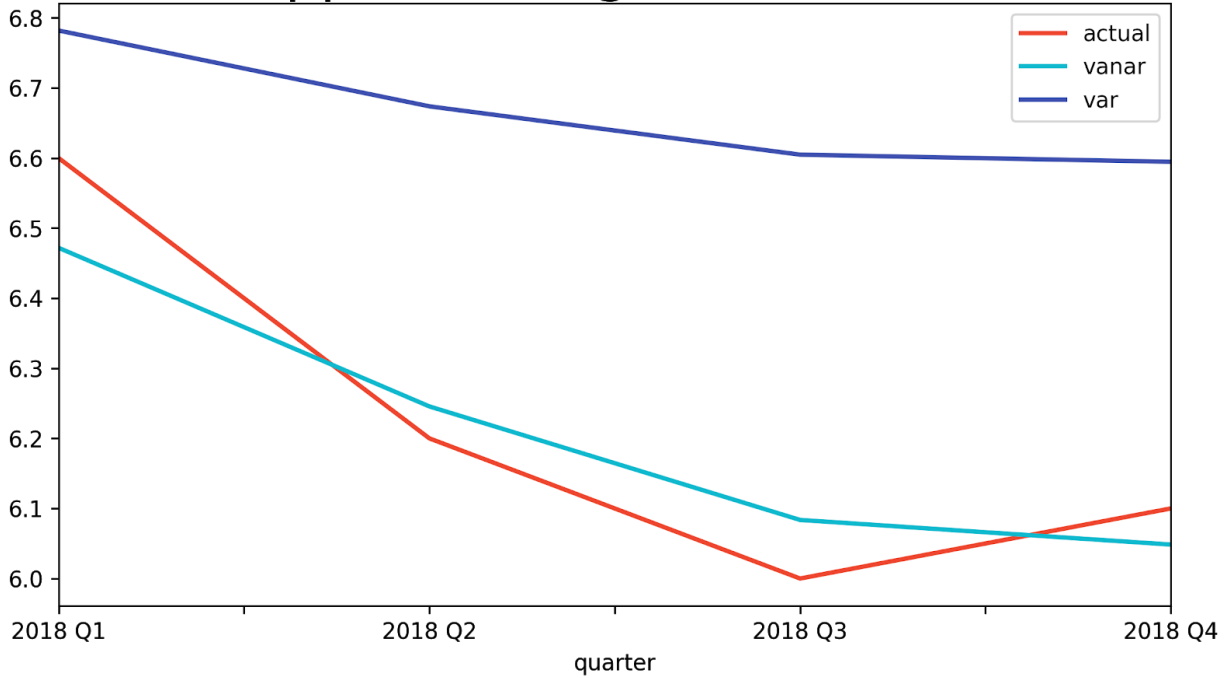

**Fig 5.** Philippine GDP Year on Year Growth forecasts of VANAR and VAR

**Causality Analysis**

What are the main drivers of GDP growth? What indicators does GDP growth tend to impact the most? To answer this, we do a pairwise causality analysis between several macroeconomic indicators and GDP growth. However, the analysis is purely between GDP and other indicators, no causality is done between the non-GDP growth indicators. The resulting causality diagram is an "all to one one to all" directed graph, or star network, where the center node is GDP growth and the weights of the arrows are the generated causality scores described in definition 1 of the Granger Causality section. The closer the causality score is to 1, the stronger the evidence for a causal relationship.

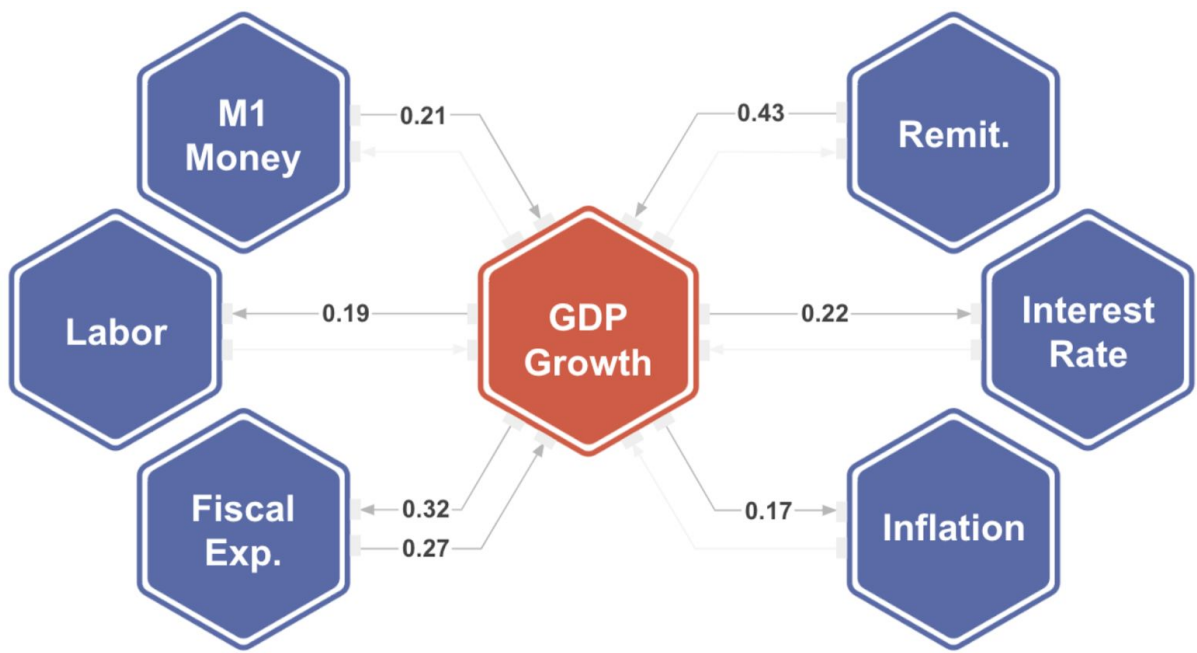

**Fig 6.** Philippines Causality Analysis

Only indicators that have positive scores (existence of a causal effect) were included as nodes for the final diagram above. Remittances has a strong, albeit one directional, causal relationship with GDP growth; it drives growth but not the other way around. Fiscal expenditure has a strong bidirectional causality with GDP growth. Growth affects interest rates but not the other way around, possibly indicating an economy driven more by fundamentals rather than movements in the financial market.

**Impulse Response**

We examine the dynamic effects of expansionary monetary policy by applying a positive shock of 0.3 to the M1 Money Supply variable during the 4th quarter of 2018, equivalent to a 30% increase in the level variable. The impulse response is generated for up to three years ahead.

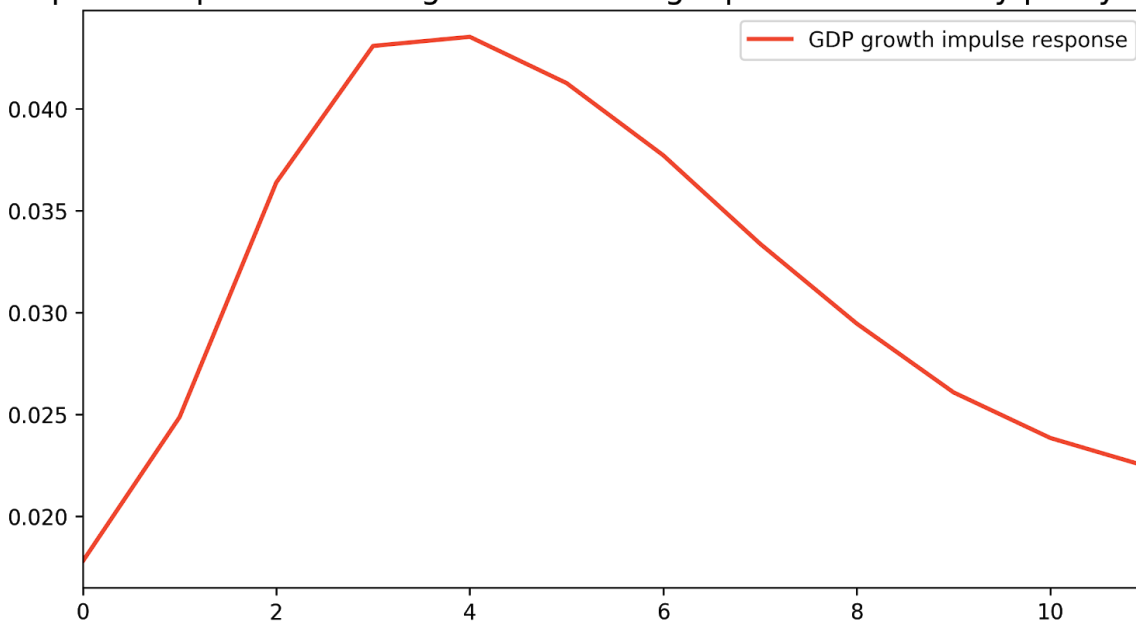

**Fig 7.** Impulse Response of GDP Growth from an Expansionary Monetary Policy Shock

The impulse response fundamentally represents the difference between predicted GDP growth with a shock and predicted GDP growth without a shock. The first period in the graph above is the period immediately after the origin point of the shock. A 30% increase in M1 money supply leads to a peak in GDP growth around a year after the shock where we see GDP growth increase by around four percentage points (if predicted GDP growth without the shock was 5.4 then the predicted GDP growth given the shock would be 5.44). The effect, however, decays after three years.

Similarly, we now examine the effects of applying shock of 0.3 to the Fiscal Expenditure variable during the 4th quarter of 2017, equivalent to a 35% increase in the level variable. Thus, instead of expansionary monetary policy, we have an increase in government spending and investments, otherwise known as expansionary fiscal policy.

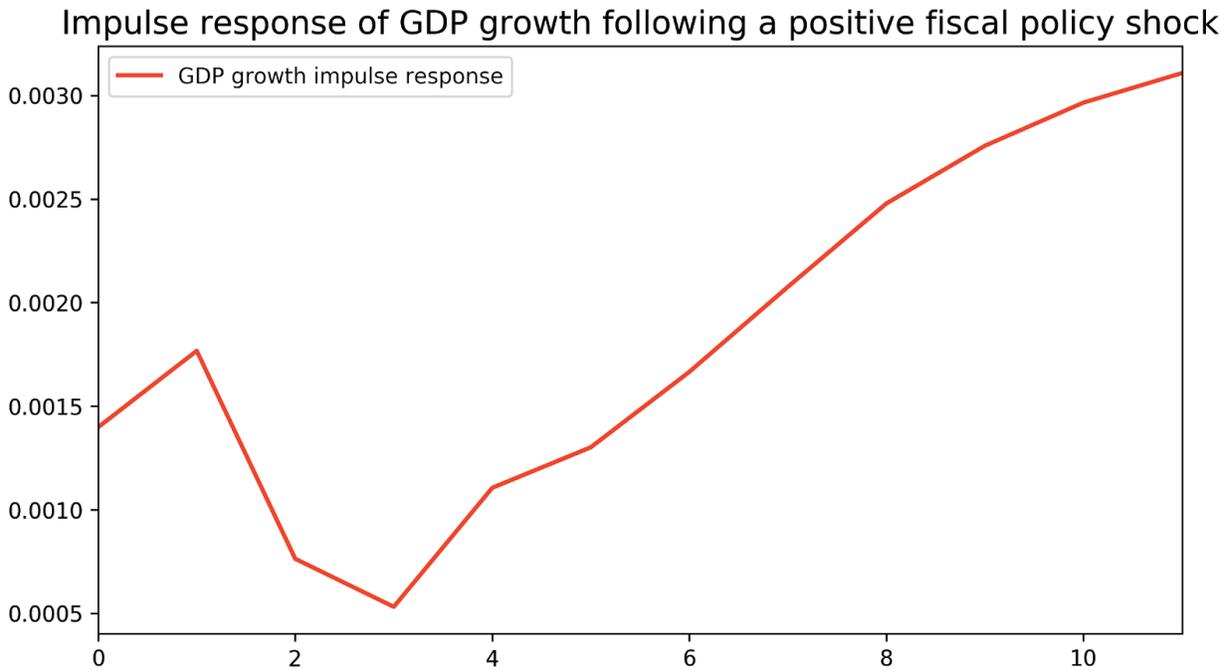

**Fig 8.** Impulse Response of GDP Growth from an Expansionary Fiscal Policy Shock

Expansionary policy has little to no effect on GDP growth in the short term but steadily increases GDP growth in the long term. This could imply the crowding out effect of fiscal policy in the short run but a positive real structural change in the long run.

**Univariate One Step Ahead Forecast Test**

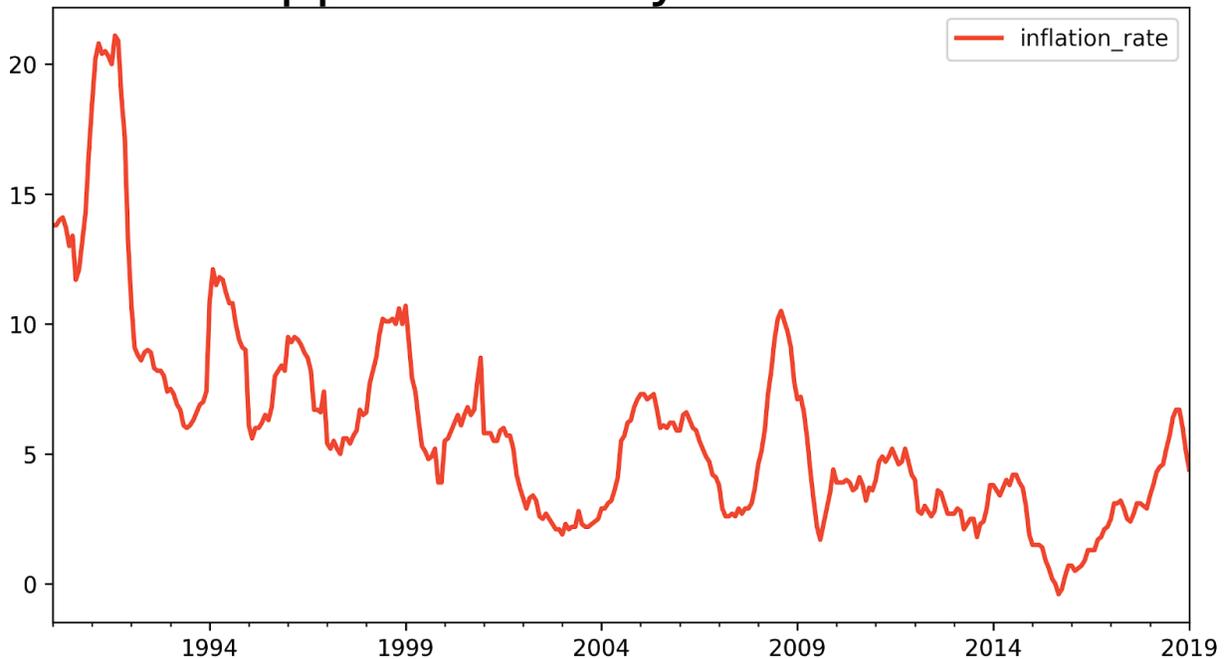

**Fig 9.** Philippine Monthly Inflation Rate time series from February 1990 to January 2019.

The error metric used here was the Root Mean Squared Scaled Error which is just the RMSE of the model divided by the RMSE of a naive forecast where a naive forecast is using the value at the current period as the forecast for the next period. Thus an RMSSE greater than one signifies a model forecast worse than a naive forecast while an RMSSE of less than one signifies true gains in forecasting accuracy. For inflation rate, ANA was the best performing model.

**Table 2.** Philippine monthly Inflation Rate one step ahead forecast RMSSE for SARIMA, TBATS, MLP, and ANA.

| Model | SARIMA | TBATS | MLP | ANA-3 Deactivated Autoencoder |
|---|---|---|---|---|
| **RMSSE** | 0.744 | 0.808 | 0.954 | 0.704 |

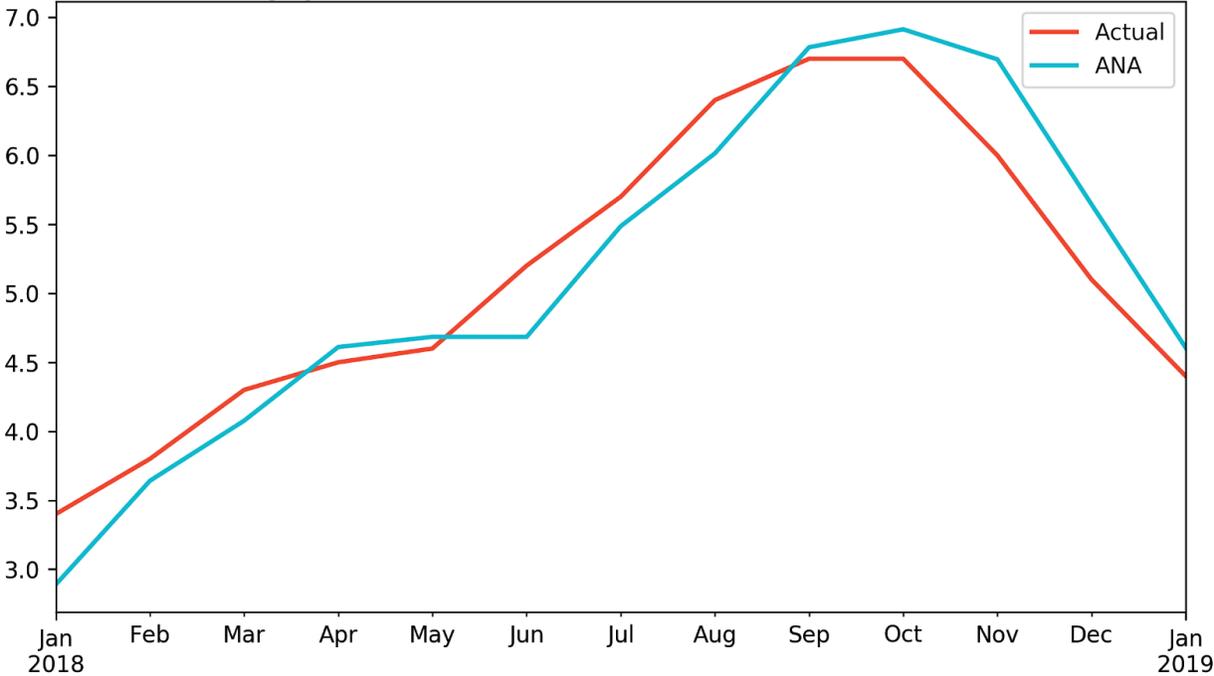

**Fig 10.** Philippine Inflation Rate test set forecasts for ANA. Test set shown above was the last 12 months of 2018 and January of 2019

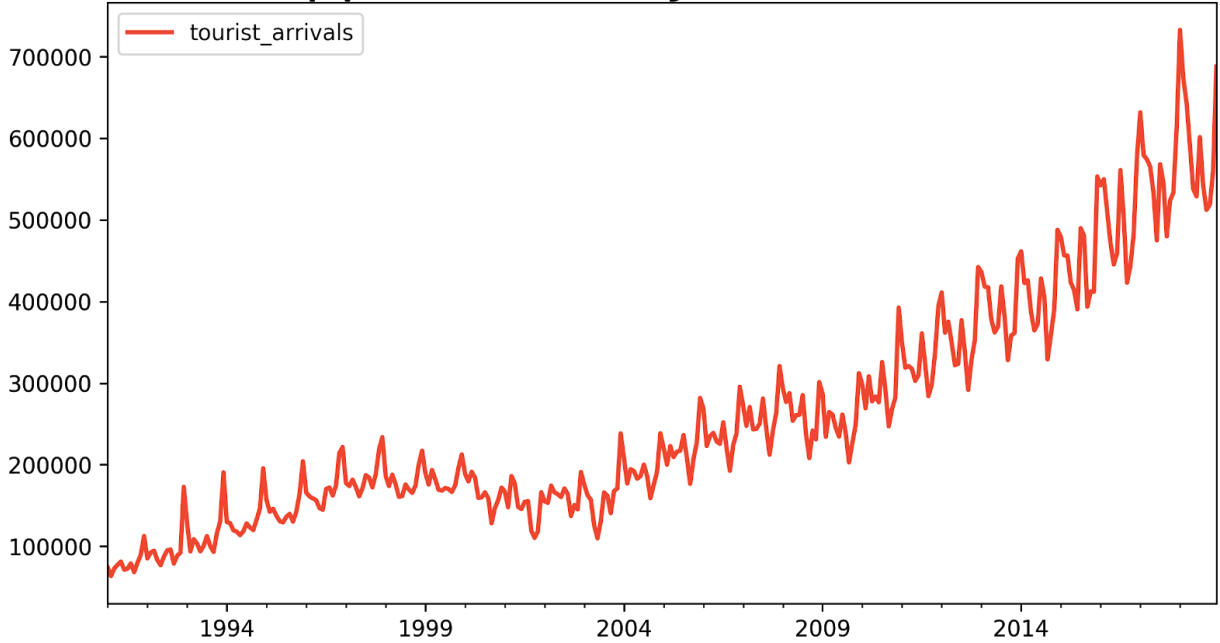

**Fig 11.** Philippine Monthly Tourist Arrivals time series from January 1991 to December 2018.

Similarly, for the Tourist Arrivals time series, ANA was the best performing model and with a considerable margin over the other models.

**Table 3.** Philippine monthly Tourist Arrivals one step ahead forecast RMSSE for SARIMA, TBATS, MLP, and ANA. Test set was comprised of 24 observations, a two year monthly range.

| Model | SARIMA | TBATS | MLP | **ANA-12 Activated Autoencoder** |
|---|---|---|---|---|
| **RMSSE** | 0.935 | 0.892 | 0.726 | 0.567 |

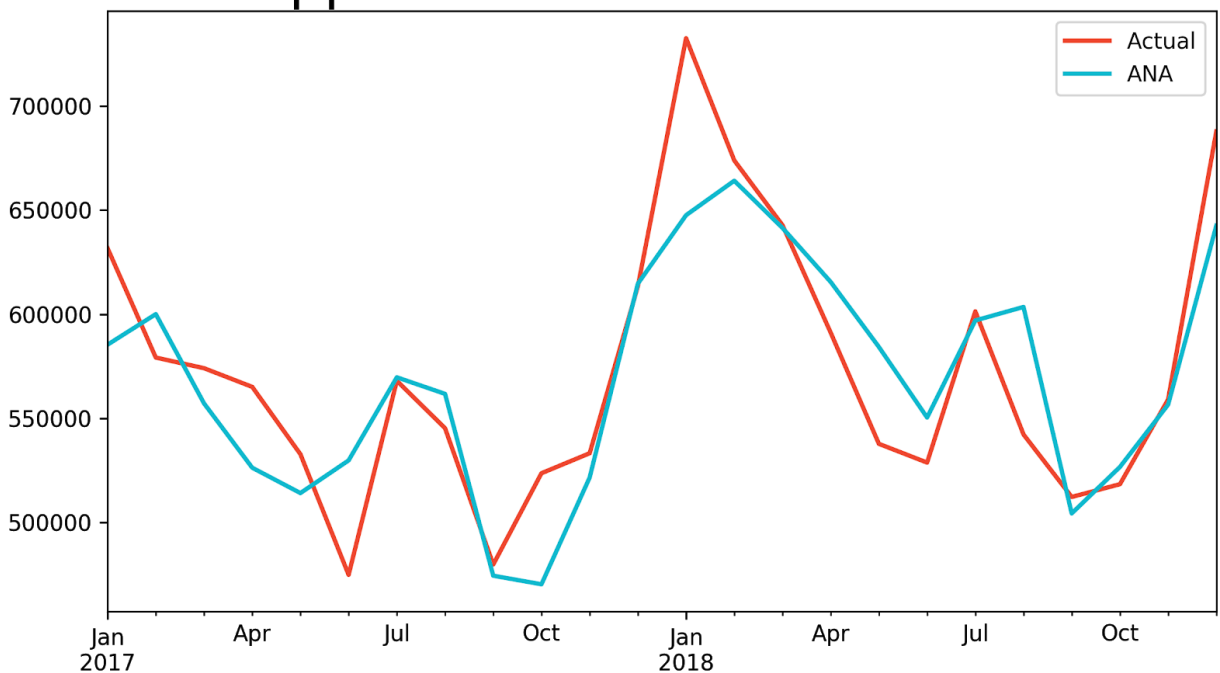

**Fig 12.** Philippine Tourist Arrivals test set forecasts for ANA. Test set shown above was 2017 and 2018

# Conclusion

Here, we demonstrated the ability of VANAR to outperform various multivariate (VAR) and univariate (SARIMA, TBATS, MLP) time series models in terms of forecasting accuracy, causality detection, and impulse response analysis. The promising performance of VANAR on our nonlinear simulated system is important, considering the prevalence of non-linearity in many real world applications [3]. Notably, we also demonstrated VANAR's robustness, still outperforming VAR even in conditions with high noise, or when training data was limited. Indeed, VANAR's performance on empirical macroeconomic data showed its promise in analysis of real-world systems, and further applications to neuroscience, ecology, medicine, and more macroeconomics remain to be explored. Given its robust performance across different datasets and different time series tasks, VANAR is a strong candidate for use in more real world dynamical systems for policymakers and researchers alike.

## **Compliance with Ethical Standards:**

Funding: This study was funded by Thinking Machines Data Science.

Conflict of Interest: The authors declare that they have no conflict of interest.
# References

[1] Banks et al. (2018). On Devaney's Definition of Chaos. The American Mathematical Monthly, 99(4)

[2] Bontempi, G., Taieb, S. B., & Le Borgne, Y. A. (2012). Machine learning strategies for time series forecasting. In European business intelligence summer school (pp. 62-77). Springer, Berlin, Heidelberg.

[3] De Gooijer, J. G., & Hyndman, R. J. (2006). 25 years of time series forecasting. International journal of forecasting, 22(3), 443-473.

[4] De Livera, A.M., Hyndman, R.J., & Snyder, R. D. (2011). Forecasting time series with complex seasonal patterns using exponential smoothing. Journal of the American Statistical Association, 106(496), 1513-1527.

[5] Fox, E., Tank, A., Covert, I., Foti, N., & Shojaie, A. (2018). Neural granger causality for nonlinear time series. arXiv:1802.05842

[6] Hornik, K., Stinchcombe, N., & White, H. (1989). Multilayer feedforward networks are universal approximators. Neural Networks, 2: 359-366

[7] Kilian, L., & Lütkepohl, H. (2017). Structural Vector Autoregressive Analysis (Themes in Modern Econometrics). Cambridge: Cambridge University Press. doi:10.1017/9781108164818
31

# APPENDIX

Here we present detailed results of VANAR (and VAR) tests across test systems, scenarios, and data richness environments.

## System 1: Two-variable nonlinear system with mirage correlation

### Default Scenario

#### High Data Richness Environment

- Training data = 850 steps

| n steps ahead for variable X | VANAR-14. Activated autoencoder. | ANA | VAR-14 | AR-14 |
| --- | --- | --- | --- | --- |
| 20 | 0.1356056 | 0.1768853 | 0.257 | 0.230 |
| 10 | 0.08673972 | 0.1174932 | 0.293 | 0.258 |

| n steps ahead for variable Y | VANAR-14. Activated autoencoder. | ANA | VAR-14 | AR-14 |
| --- | --- | --- | --- | --- |
| 20 | 0.01805058 | 0.04122844 | 0.064 | 0.054 |
| 10 | 0.00580254 | 0.04659928 | 0.060 | 0.037 |

#### Medium Data Richness Environment

- Training data = 250 steps

| n steps ahead for variable X | VANAR-9. Activated autoencoder. | ANA | VAR-9 | AR-9 |
|---|---|---|---|---|
| 20 | 0.1196829 | 0.130279 | 0.206 | 0.242 |
| 10 | 0.1060848 | 0.1207651 | 0.202 | 0.238 |

| n steps ahead for variable Y | VANAR-9. Activated autoencoder. | ANA | VAR-9 | AR-9 |
|---|---|---|---|---|
| 20 | 0.01846403 | 0.02132268 | 0.020 | 0.018 |
| 10 | 0.01318353 | 0.01881621 | 0.016 | 0.016 |

## Low Data Richness Environment

- Training data = 50 steps

| n steps ahead for variable X | VANAR-14. Activated autoencoder. | ANA | VAR-4 | AR-4 |
|---|---|---|---|---|
| 20 | 0.2471158 | 0.3300185 | 0.265 | 0.245 |
| 10 | 0.2415083 | 0.12016 | 0.245 | 0.216 |

| n steps ahead for variable Y | VANAR-14. Activated autoencoder. | ANA | VAR-4 | AR-4 |
|---|---|---|---|---|
| 20 | 0.03056499 | 0.0373002 | 0.031 | 0.084 |
| 10 | 0.02755299 | 0.04316465 | 0.035 | 0.085 |

# No Interaction Scenario

We repeat the methods in the Default Scenario but with a different dataset: we simulate System 1 where the two variables have no interaction (i.e., $Y_{(t-1)}$ no longer appears in the formula for $X_{(t)}$, and vice-versa).

## High Data Richness Environment

- Training data = 850 steps

| n steps ahead for variable X | VANAR-65 Deactivated autoencoder. | ANA | VAR-65 | AR-65 |
|---|---|---|---|---|
| 20 | 0.2206835 | 0.2058791 | 0.233 | 0.232 |
| 10 | 0.1757822 | 0.1546297 | 0.201 | 0.189 |

| n steps ahead for variable Y | VANAR-65 Deactivated autoencoder. | ANA | VAR-65 | AR-65 |
|---|---|---|---|---|
| 20 | 0.01711407 | 1.744163e-05 | 0.090 | < 0.001 |
| 10 | 0.01885756 | 1.447229e-05 | 0.090 | < 0.001 |

## Medium Data Richness Environment

- Training data = 250 steps

| n steps ahead for variable X | VANAR-24, Deactivated autoencoder. | ANA | VAR-24 | AR-24 |
|---|---|---|---|---|
| 20 | 0.2662899 | 0.2304825 | 0.300 | 0.243 |
| 10 | 0.233162 | 0.1979543 | 0.358 | 0.252 |

| n steps ahead | VANAR-24 | ANA | VAR-24 | AR-24 |
|---|---|---|---|---|

| for variable Y | Deactivated autoencoder. | | | |
|---|---|---|---|---|
| 20 | 0.01255132 | 0.0003848478 | 0.090 | <0.001 |
| 10 | 0.01101204 | 0.0002955276 | 0.090 | <0.001 |

## Low Data Richness Environment

- Training data = 50 steps

| n steps ahead for variable X | VANAR-5 Deactivated autoencoder. | ANA | VAR-5 | AR-5 |
|---|---|---|---|---|
| 20 | 0.2703183 | 0.2588373 | 0.272 | 0.245 |
| 10 | 0.2548257 | 0.2326701 | 0.276 | 0.225 |

| n steps ahead for variable Y | VANAR-5 Deactivated autoencoder. | ANA | VAR-5 | AR-5 |
|---|---|---|---|---|
| 20 | 0.1077617 | 0.003335119 | 0.095 | 0.008 |
| 10 | 0.1162037 | 0.002734291 | 0.093 | 0.004 |

# With Observation Noise 1 Scenario

We repeat the methods in the Default Scenario but with a different dataset: we simulate System 1 but with observation noise. Noise is introduced at each time step as a normally-distributed random variable with mean = 0 and sd = 0.1.

## High Data Richness Environment

- Training data = 850 steps

| n steps ahead | VANAR-25. | ANA | VAR-25 | AR-25 |
|---|---|---|---|---|

| for variable X | Activated autoencoder | | | |
|---|---|---|---|---|
| 20 | 0.2450847 | 0.2475296 | 0.310 | 0.260 |
| 10 | 0.27111667 | 0.2804906 | 0.365 | 0.295 |

| n steps ahead for variable Y | VANAR-25. Activated autoencoder | ANA | VAR-25 | AR-25 |
|---|---|---|---|---|
| 20 | 0.09659208 | 0.04982226 | 0.140 | 0.104 |
| 10 | 0.1021324 | 0.02934832 | 0.173 | 0.113 |

## Medium Data Richness Environment

- Training data = 250 steps

| n steps ahead for variable X | VANAR-9. Activated autoencoder. | ANA | VAR-9 | AR-9 |
|---|---|---|---|---|
| 20 | 0.04789752 | 0.2531016 | 0.261 | 0.262 |
| 10 | 0.03368176 | 0.07449114 | 0.292 | 0.271 |

| n steps ahead for variable Y | VANAR-9 Activated autoencoder. | ANA | VAR-9 | AR-9 |
|---|---|---|---|---|
| 20 | 0.01743703 | 0.04430856 | 0.131 | 0.104 |
| 10 | 0.01130574 | 0.03508665 | 0.125 | 0.098 |

## Low Data Richness Environment

- Training data = 50 steps

| n steps ahead for variable X | VANAR-5. Activated autoencoder. | ANA | VAR-5 | AR-5 |
|---|---|---|---|---|
| 20 | 0.2593454 | 0.225887 | 0.303 | 0.310 |
| 10 | 0.2452869 | 0.2311914 | 0.212 | 0.263 |

| n steps ahead for variable Y | VANAR-5. Activated autoencoder. | ANA | VAR-5 | AR-5 |
|---|---|---|---|---|
| 20 | 0.1071115 | 0.1549625 | 0.162 | 0.129 |
| 10 | 0.1141862 | 0.1758735 | 0.180 | 0.155 |

## With Observation Noise 2 Scenario

We repeat the methods in the Observation Noise 1 Scenario but with the noise term having a standard deviation of 0.01.

### High Data Richness Environment

- Training data = 850 steps

| n steps ahead for variable X | VANAR-14. Activated autoencoder. | ANA | VAR-14 | AR-14 |
|---|---|---|---|---|
| 20 | 0.1770601 | 0.3436085 | 0.257 | 0.231 |
| 10 | 0.08852866 | 0.2923459 | 0.298 | 0.260 |

| n steps ahead for variable Y | VANAR-14. Activated autoencoder. | ANA | VAR-14 | AR-14 |
|---|---|---|---|---|
| 20 | 0.0193065 | 0.03161021 | 0.053 | 0.042 |
| 10 | 0.009916833 | 0.03103254 | 0.063 | 0.045 |

## Medium Data Richness Environment

- Training data = 250 steps

| n steps ahead for variable X | VANAR-9. Activated autoencoder. | ANA | VAR-9 | AR-9 |
|---|---|---|---|---|
| 20 | 0.2541653 | 0.4106635 | 0.210 | 0.246 |
| 10 | 0.1450822 | 0.3488136 | 0.200 | 0.239 |

| n steps ahead for variable Y | VANAR-9. Activated autoencoder. | ANA | VAR-9 | AR-9 |
|---|---|---|---|---|
| 20 | 0.02026843 | 0.02010566 | 0.022 | 0.022 |
| 10 | 0.009910785 | 0.01341163 | 0.018 | 0.016 |

## Low Data Richness Environment

- Training data = 50 steps

| n steps ahead for variable X | VANAR-5. Activated autoencoder. | ANA | VAR-5 | AR-5 |
|---|---|---|---|---|
| 20 | 0.2132341 | 0.2160254 | 0.266 | 0.243 |
| 10 | 0.2275326 | 0.1893417 | 0.249 | 0.209 |

| n steps ahead for variable Y | VANAR-5. Activated autoencoder. | ANA | VAR-5 | AR-5 |
|---:|---|---:|---:|---:|
| 20 | 0.02770695 | 0.04308755 | 0.030 | 0.077 |
| 10 | 0.02502873 | 0.05695597 | 0.035 | 0.081 |